\documentclass[lettersize,journal]{IEEEtran}

\usepackage{amsmath,amsfonts}
\usepackage{algorithm}
\usepackage{array}
\usepackage[caption=false,font=normalsize,labelfont=sf,textfont=sf]{subfig}
\usepackage{textcomp}
\usepackage{stfloats}
\usepackage{url}
\usepackage{verbatim}
\usepackage{graphicx}
\usepackage{cite}
\IEEEoverridecommandlockouts
\usepackage{amssymb}
\usepackage{xcolor}
\usepackage{color,soul}
\usepackage{caption}
\usepackage{subcaption}
\usepackage{booktabs}
\usepackage{tabularx}
\usepackage{multirow}
\usepackage{algpseudocode}

\usepackage{siunitx} 

\usepackage{pifont}

\usepackage{xspace}
\newcommand{\model}{{\textsc{Cefira}}\xspace}

\begin{document}

\title{\model: Modular Correlation-Aware Ensemble for VNF Performance Forecasting in Computing Continuum}

\author{
Milad~Rafiee,
Andres~Ocampo-Palacio,
Amir~Taherkordi,~\IEEEmembership{Senior Member,~IEEE,}
and
\"{O}zg\"{u}~Alay,~\IEEEmembership{Senior Member,~IEEE}
\thanks{Milad Rafiee, Amir Taherkordi, and \"{O}zg\"{u} Alay are with the Department of Informatics, University of Oslo, Oslo, Norway. \"{O}zg\"{u} Alay is also with the Computer Science Department, Karlstad University, Karlstad, Sweden.}
\thanks{Andres Ocampo-Palacio is with the Department of Sustainable Communication Technologies, SINTEF Digital, Oslo, Norway.}
\thanks{Corresponding author: Milad Rafiee (e-mail: miladraf@ifi.uio.no).}
}

\maketitle

\begin{abstract}

Accurate forecasting of Virtual Network Function (VNF) performance indicators is essential for proactive resource management in Network Function Virtualization (NFV) environments. However, existing approaches typically model each VNF as an isolated time series and treat performance metrics as independent channels, and assume failure-free operation. 
These three simplifications limit their applicability to real-world Service Function Chains (SFCs).
This paper proposes \model, a VNF resource demand forecasting model that jointly captures Intra-VNF performance metrics correlation, Inter-VNF topological dependency, and heterogeneous temporal dynamics. \model is composed of three dedicated modules. A residual module learns Intra-VNF correlations within each VNF. A topology-aware gated message-passing module propagates latent VNF states across the SFC graph while gating on a binary crash indicator to suppress corrupted signals from crashing or recovering neighbors. Finally, a heterogeneous temporal forecasting module produces the final forecast.
The model is evaluated on a containerized SFC testbed comprising two different topologies under non-shared and shared-VNF configurations.
The evaluation results show that \model achieves an average $R^2$ of up to 0.98 and 
crash-aware gating achieves error reductions of up to 22\% on high-crash-rate VNFs.
\model achieves measurable and consistent gains in forecasting accuracy, supporting reliable resource management in real-world and large-scale cloud-native SFC deployments.

\end{abstract}

\begin{IEEEkeywords}
Virtual Network Function (VNF), Service Function Chain (SFC), Performance Metrics Forecasting, Computing Continuum
\end{IEEEkeywords}

\vspace{-5pt}
\section{Introduction}
\label{sec:introduction}

Cloud-native network services are increasingly deployed across distributed computing infrastructures. In such environments, traffic is steered through an ordered sequence of Virtual Network Functions (VNFs), known as a Service Function Chain (SFC), which may include firewall, intrusion detection, and traffic inspection functions~\cite{wu2024graph}. 
These functions process workloads generated by diverse applications, including IoT deployments, smart-city services, and connected vehicles. Their resource demands vary with traffic intensity, service characteristics, and network and operating conditions~\cite{nguyen2019placement, zhang2021multi, de2023optimal}.

In such environments, reactive resource allocation based solely on current measurements often leads to over-provisioning, which wastes infrastructure capacity, or under-provisioning, which degrades service quality and may trigger service disruption~\cite{yi2024traffic}. Proactive resource management addresses these limitations by relying on accurate ahead-of-time forecasting of VNF resource demands to inform auto-scaling, placement, and orchestration decisions within NFV-MANO~\cite{etsi_nfv_mano}.

Early studies on VNF resource demand prediction focused on forecasting a single performance metric, most commonly CPU utilization or traffic load~\cite{luo2019scaling, schneider2020machine, subramanya2021centralized, cho2021deep, yi2024traffic}. However, traffic load, CPU, and memory are inherently interdependent, making single-metric models structurally insufficient for capturing future resource requirements. 
Multivariate forecasting approaches that jointly model multiple performance metrics have since been proposed~\cite{kim2019deep, rafiee2025proactive}, but most still treat VNFs as independent entities, ignoring that workload changes propagate between neighboring functions through traffic flows. 
This has motivated graph-based learning methods that exploit SFC topology to model spatial relationships among VNFs~\cite{wu2023graph, wu2024graph, xiao2025st, mijumbi2017topology, zhang2023forecast}.

Despite these advances, existing methods remain limited along three dimensions. 
First, existing methods typically rely on a single temporal architecture, or combine at most one or two, such as LSTM, CNN-LSTM, or a graph neural network~\cite{kim2019deep, subramanya2021centralized, cho2021deep, zhang2023forecast}. Since VNF telemetry is non-stationary~\cite{wu2024graph} and the best-fitting architecture can vary across VNFs and over time, committing to one architecture in advance risks a poor fit in some conditions; an ensemble instead combines several architectures and weights their contributions per forecast.
Second, none of the topology-aware approaches reviewed above account for VNF failures or crash events, and are developed and evaluated under an implicit assumption of stable, failure-free operation.

In practice, cloud-native VNFs are commonly deployed as containers orchestrated by platforms such as Kubernetes~\cite{Kubernetesframework}, and frequently experience crashes triggered by Out-of-Memory (OOM) kills, liveness probe failures, or resource limit violations~\cite{di2026reliability}. Such events produce abrupt discontinuities in performance trajectories, CPU and memory usage drop to near zero or traffic collapses, followed by a recovery curve as Kubernetes restarts the container. Without explicit failure information, a model cannot distinguish low traffic due to normal low load from low traffic due to an upstream crash, and the resulting corrupted signal propagates downstream through the SFC. No existing work addresses this ambiguity.

Third, no existing method combines multivariate metric modeling, topology awareness, and diverse temporal modeling in a single model: even methods that jointly model multiple metrics and topology~\cite{mijumbi2017topology, zhang2023forecast} address only two of these three.

Motivated by these limitations, this paper proposes \model (\underline{c}orrelation-aware \underline{e}nsemble for VN\underline{F} performance metr\underline{i}cs fo\underline{r}ec\underline{a}sting), a unified forecasting model that addresses them simultaneously. 
\model\ has three modules, one for each dimension above. The Intra-VNF module captures linear and non-linear correlations among performance metrics within each VNF. The Inter-VNF module models interactions among VNFs along the directed SFC structure, with crash-aware message gating to suppress corrupted signals from failing neighbors. The temporal ensemble integrates RNN, LSTM, GRU, and Transformer backbones through stacking-based meta-learning, so that no single architecture has to fit every VNF and operating condition.
A binary crash indicator is incorporated as an explicit input feature. The indicator conditions the Intra-VNF latent representation before spatial processing, gates inter-VNF message passing to suppress corrupted signals from crashing neighbors, and informs the temporal backbones to learn crash-conditioned dynamics.

The main contributions of this work are:
\begin{itemize}
    \item \model: a modular VNF forecasting model that jointly capturing Intra-VNF correlation, Inter-VNF topology, and heterogeneous temporal dynamics, forecasting CPU, memory, and traffic load simultaneously across all SFC nodes.
    \item An Intra-VNF residual module capturing linear and non-linear metric correlations within each VNF.
    \item A topology- and crash-aware graph propagation module that gates inter-VNF message passing to suppress failure-induced signal corruption.
    \item A heterogeneous temporal ensemble combining RNN, LSTM, GRU, and Transformer backbones via stacking-based meta-learning, which adapts to varying temporal behavior across VNFs and operating conditions.
    \item A cloud-native SFC testbed of containerized VNFs orchestrated through Kubernetes, which enables the evaluation of \model using IoT-based workloads spanning four representative deployment scales and real 4G/5G uplink measurements. The testbed and dataset are publicly released to support reproducibility (Section~\ref{sec:experimental_setup}).
\end{itemize}

This paper is organized as follows. Section
\ref{sec:related_work} reviews related work on VNF resource prediction. Section
\ref{sec:high_level_architecture} presents the high-level system architecture. Section
\ref{sec:forecasting_model} details the proposed \model framework. Section
\ref{sec:experimental_setup} describes the experimental setup and evaluation methodology. Section
\ref{sec:evaluationresult} presents and discusses the evaluation results. Section
\ref{sec:conclusion} concludes the paper and highlights directions for future research.

\section{Related Work}
\label{sec:related_work}

Extensive research has investigated resource forecasting for VNF in NFV environments, which can be broadly classified into univariate and multivariate forecasting approaches, as well as model-based and topology-aware methods.

Early studies mainly focus on univariate prediction, typically targeting CPU utilization or traffic load. Luo et al. \cite{luo2019scaling} combine RNN-based traffic forecasting with a reinforcement learning agent for VNF chain scaling and placement, optimizing operational and deployment costs. Schneider et al. \cite{schneider2020machine} evaluate multiple regression models on VNF profiling data to estimate CPU demand and integrate the predictions into placement decisions, showing that accurate forecasting reduces both over- and under-provisioning. Subramanya and Riggio \cite{subramanya2021centralized} explore deep learning models for traffic load forecasting and scaling decisions, including LSTM and CNN-LSTM variants, and further introduce a federated learning approach for distributed training across edge nodes.

To capture dependencies in service chains, several works incorporate spatial or graph-based structures. Cho et al. \cite{cho2021deep} propose a spatiotemporal model combining LSTM and CNN-LSTM to exploit correlations between upstream and downstream VNFs for forecasting CPU usage, although SFC topology is not explicitly modeled. Wu et al. \cite{wu2023graph} introduce a graph attention-based model for atom-level VNFs, capturing hierarchical dependencies within service chains for traffic load forecasting. Wu et al. \cite{wu2024graph} further extend this direction using a graph transformer with temporal decoding to model multi-step workload forecasting in SFCs, showing that topology-aware modeling reduces prediction uncertainty. Xiao et al. \cite{xiao2025st} propose a spatio-temporal architecture integrating graph convolutional attention with state-space sequence modeling to improve long-horizon VNF CPU usage forecasting.

In parallel, multivariate forecasting has gained attention to jointly model multiple performance metrics such as CPU, memory, and latency. Kim et al. \cite{kim2019deep} propose CAT-LSTM with multivariate SFC-wide resource history (CPU, memory, disk, process count) to predict a target VNF's CPU utilization, showing that inter-VNF dependencies improve univariate prediction. Mijumbi et al. \cite{mijumbi2017topology} propose a graph-based formulation for multivariate VNF resource prediction, leveraging topology information to model inter-node influence and jointly forecasting CPU, memory usage, and processing delay, although their approach is limited to single-step forecasting and synthetic traffic traces. Zhang et al. \cite{zhang2023forecast} integrate graph neural networks with multitask learning to jointly forecast CPU, storage, and processing latency and incorporate the results into SFC placement optimization.

Existing studies mostly commit to a single temporal architecture and treat topology-aware and multivariate modeling separately, which limits their ability to capture the full complexity of real-world SFC environments. VNF resource metrics are known to co-evolve under shared workloads, as CPU, memory, and traffic load reflect the same underlying service demand~\cite{zhang2023forecast}.
In our prior work~\cite{rafiee2025proactive}, we proposed a stacked ensemble combining RNN, LSTM, GRU, and Transformer models for multivariate VNF forecasting, improving robustness across diverse traffic patterns. However, that work does not model SFC topology, Inter-VNF dependencies, or crash states, which the present paper addresses.

\section{System Architecture}
\label{sec:high_level_architecture}

This section presents the overall system architecture in which the proposed VNF performance forecasting framework operates, as illustrated in Figure~\ref{fig: High Level Architecture}. The architecture comprises four components: the Traffic and Workload Generation layer, the RAN/Mobile Core, and the Computing Continuum, and NFV-MANO, which hosts the proposed framework.

Traffic originates at the Traffic and Workload Generation layer, traverses the mobile network (RAN and mobile core), and reaches the computing continuum, where it is processed by an ordered chain of VNFs, running on the virtualized compute, storage, and network resources that host them. These VNFs and their hosting infrastructure correspond to two of the three main domains of the ETSI NFV architectural framework, VNFs and NFV Infrastructure (NFVI), the third being NFV-MANO~\cite{etsi_nfv002}. The management, monitoring, and forecasting functions of the proposed framework are detailed in Section~\ref{nfv-mano}, which covers its functional blocks and the placement of the proposed framework within them.

\vspace{-5pt}
\subsection{Traffic and Workload Generation}

The Traffic and Workload Generation Layer represents the sources of traffic and service requests processed by the SFC. It abstracts different workload producers that generate traffic streams delivered to the communication infrastructure. The characteristics of these workloads, including traffic intensity, request frequency, and data volume, influence the resource consumption and operational behavior of VNFs deployed along the SFC~\cite{Mijumbi2016NFV,kuo2018deploying}.

\vspace{-5pt}
\subsection{RAN / Mobile Core}
Before reaching the computing continuum, traffic traverses the \textit{RAN / Mobile Core} segment, which represents the communication infrastructure responsible for transporting workload traffic toward the processing layer. 
RAN facilitates wireless communication between workload sources and the mobile network, supporting current cellular network deployments such as 4G and 5G technologies.
The Mobile Core then routes the data to the computing continuum, where VNF processing takes place. Together, these components define the communication path and traffic conditions under which the VNFs, the focus of the proposed framework, operate.

\vspace{-5pt}
\subsection{Computing Continuum}
\begin{figure}[htbp]
    \centering
    \includegraphics[width=12cm, height=7.5cm]{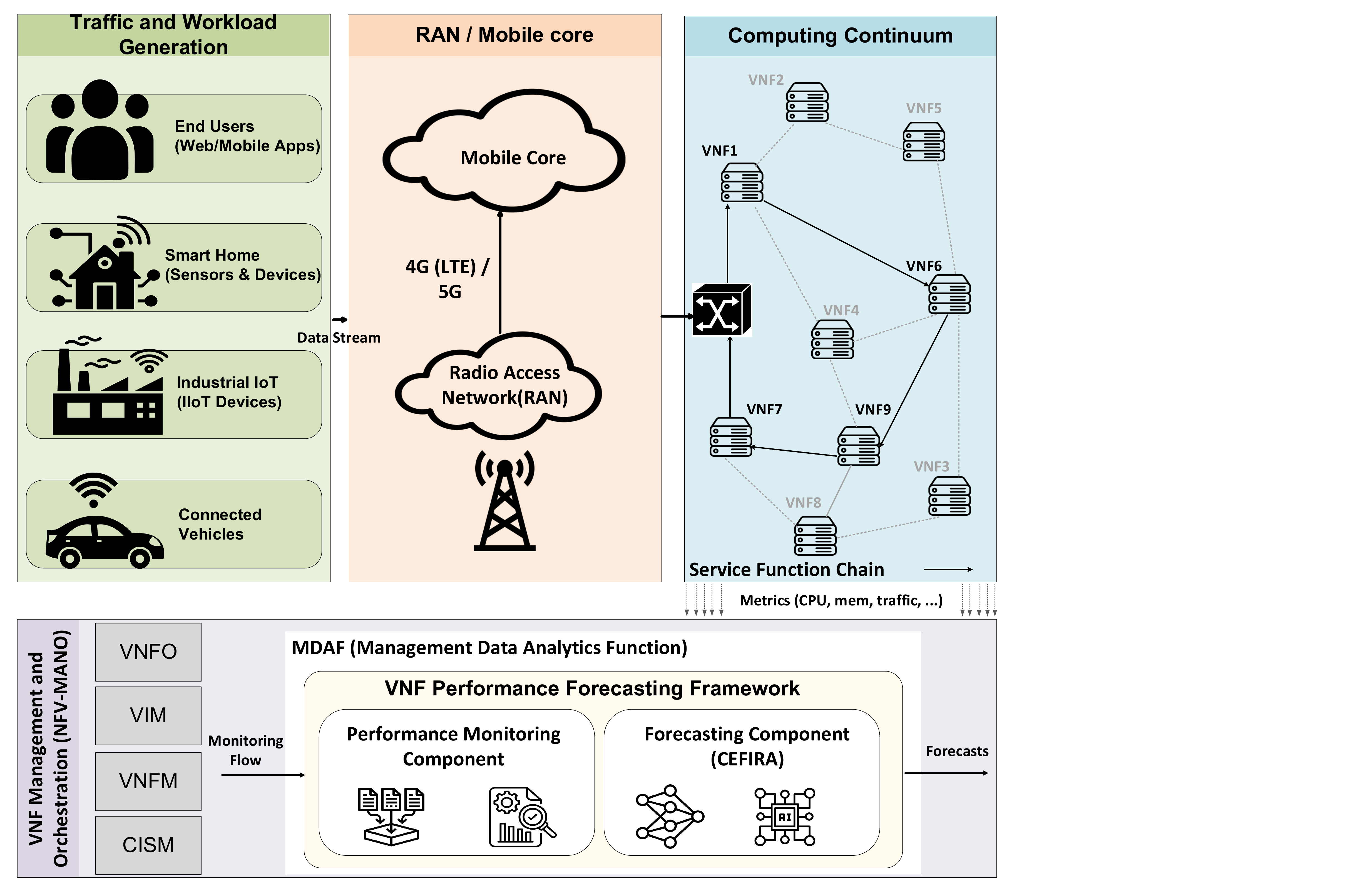}
    \caption{System architecture: traffic path from workload sources through the RAN/Mobile Core to the computing continuum, and NFV-MANO hosting the VNF Performance Forecasting Framework.}
    \label{fig: High Level Architecture}
    \vspace{-10pt}
\end{figure}

At the \textit{computing continuum}, incoming traffic is processed by an ordered chain of VNFs organized as an SFC, running on the underlying NFV Infrastructure (NFVI). As depicted in Figure~\ref{fig: High Level Architecture}, the SFC consists of multiple VNF instances (e.g., VNF1 through VNF9) that can be flexibly composed according to application requirements.

\vspace{-5pt}
\subsection{NFV Management and Orchestration (NFV-MANO)}
\label{nfv-mano}

NFV-MANO~\cite{etsi_nfv_mano} hosts the standard functional blocks NFV Orchestrator (NFVO), VNF Manager (VNFM), and Virtualised Infrastructure Manager (VIM), together with, for the container-based deployments considered in this work, the Container Infrastructure Service Management (CISM) function~\cite{etsi_nfv_ifa040}. Beyond these, NFV-MANO hosts a Management Data Analytics Function (MDAF)~\cite{etsi_nfv_ifa041, etsi_nfv_ifa047}, responsible for processing and analysing management data collected from NFV-MANO entities to provide analytics outputs to a consumer, which can be the NFVO itself or an external Operations Support System (OSS) entity.

The proposed VNF Performance Forecasting Framework is realized as an MDAF-type analytics function within NFV-MANO, following the general producer/consumer interaction pattern defined for the MDAF. The consumer of the framework's output is an orchestrator or autoscaler, occupying the consumer role played by the NFVO or an OSS entity; the design of such a consumer is outside the scope of this paper, which focuses exclusively on the accuracy of the forecasts themselves.

The framework consists of two components. The Performance Monitoring Component continuously monitors performance metrics of the VNFs. The Forecasting Component (\model) takes this monitored data as input and produces simultaneous next-step forecasts of CPU usage, memory usage, and traffic load for every VNF in the SFC.

\subsubsection{Performance Monitoring Component}

The Performance Monitoring Component continuously monitors all VNFs within the SFC and records four metrics per VNF, at a common sampling interval detailed in Section~\ref{sec:experimental_setup}: CPU usage, memory usage, traffic load, and crash status. These four metrics are selected to directly support the forecasting objective introduced in Section~\ref{sec:introduction}: CPU and memory usage characterize the VNF resource state that \model forecasts, traffic load captures the exogenous workload that drives that resource state, and crash status disambiguates genuine failure events from normal low-load operation, addressing the crash-ambiguity problem motivating this work. 
As introduced in Section~\ref{sec:introduction}, VNFs in this work are deployed as containers orchestrated by Kubernetes; the metric definitions below reflect this assumption.

\subsubsection*{Resource Usage Metrics}
\begin{itemize}

\item \textbf{CPU Usage:} 
When CPU usage exceeds the allocated limit, Kubernetes throttles the container, producing a saturation plateau in the CPU trace. Forecasting CPU usage ahead of time is therefore valuable for proactive resource management, as it can inform decisions aimed at avoiding saturation before it occurs.

\item \textbf{Memory Usage:} 
Unlike CPU, memory is non-compressible under Kubernetes: when memory usage approaches the allocated memory, the container may be terminated with an OOMKilled signal, producing an abrupt drop to zero in both memory and CPU traces followed by a restart recovery curve\cite{di2026reliability}. 

\end{itemize}

\subsubsection*{Network Metrics}
\begin{itemize}
\item \textbf{Traffic Load:} 
Traffic load is reported by each VNF container and often serves as the primary exogenous driver of CPU and memory demand; changes in traffic volume can propagate through the SFC, affecting resource consumption at downstream VNFs.
\end{itemize}

\subsubsection*{Crash Indicator}
\begin{itemize}
\item \textbf{Crash Status:} 
The crash status is derived directly from Kubernetes pod restart events. Subsequently, this label  enters \model as the explicit crash-indicator feature, allowing \model to distinguish resource-usage patterns caused by low load from those caused by crash-and-recovery transients.
\end{itemize}

\begin{figure}[htbp]
    \centering
    \includegraphics[width=8cm, height=4.5cm]{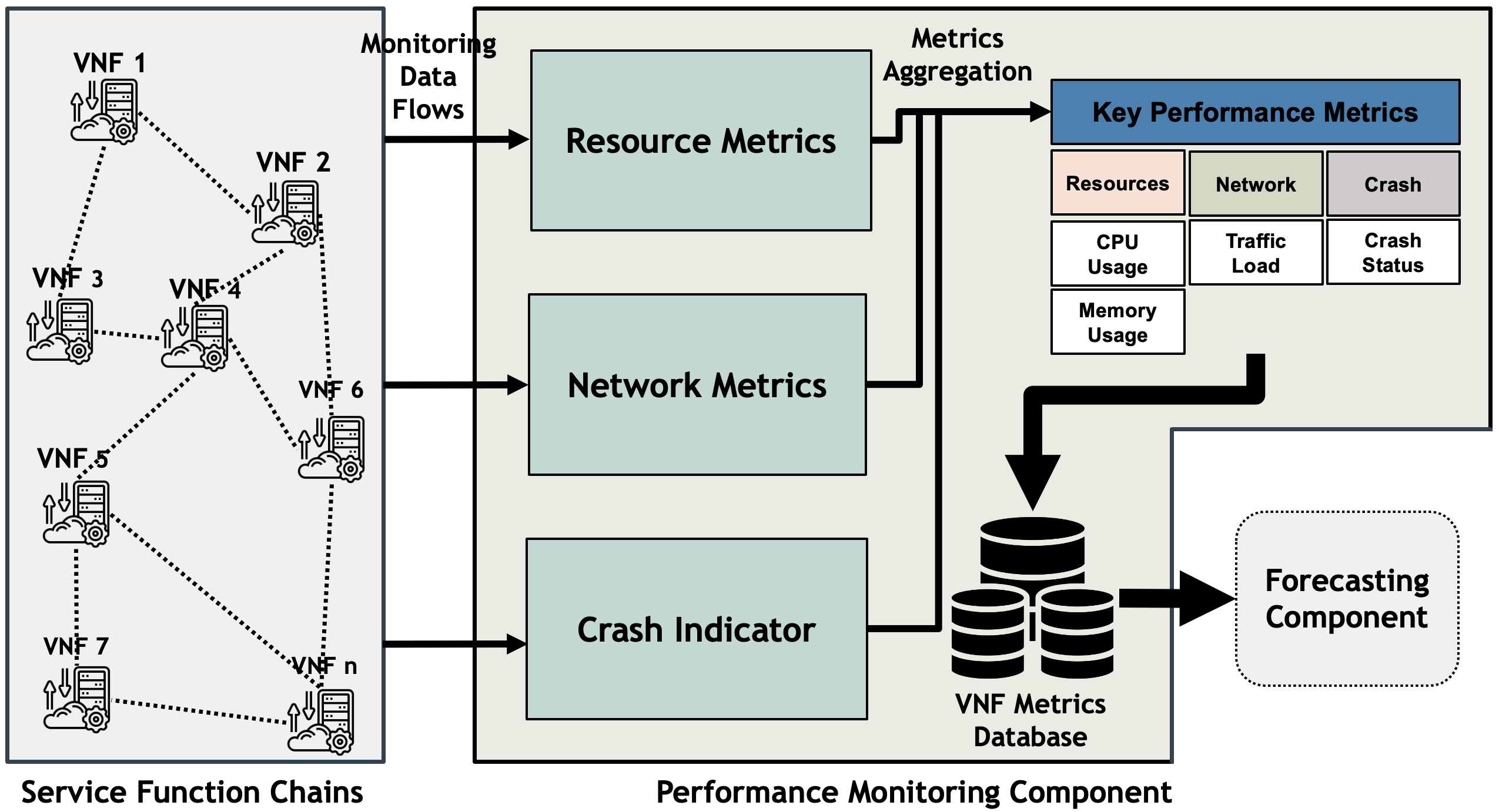}
    \caption{Performance Monitoring Component. Dedicated modules monitor CPU usage, memory usage, traffic load, and crash status from each VNF.}
    \label{fig: metrics collector}
\end{figure}

The Performance Monitoring Component is organized around a set of specialized monitoring modules, as illustrated in Figure~\ref{fig: metrics collector}. 
Because the four performance metrics originate from heterogeneous sources, CPU and memory usage are obtained from the Kubernetes monitoring infrastructure, traffic load is reported by the VNF containers themselves, and crash status is derived from Kubernetes pod restart events, dedicated modules are employed for each metric. 
Collected records are time-stamped and associated with the corresponding VNF name, making it possible to obtain a complete historical time series of each VNF’s performance metrics. These modules continuously monitor the VNFs and store the collected data in a centralized database for further processing and analysis.

\subsubsection{Forecasting Component (\model)}
\model receives the time series of monitored metrics of each VNF and produces simultaneous next-step forecasts of CPU usage, memory usage, and traffic load for every VNF in the SFC. A detailed description of its architecture is provided in Section~\ref{sec:forecasting_model}.

\section{\model}
\label{sec:forecasting_model}

\begin{figure*}[htbp]
    \centering
    \includegraphics[width=18cm, height=7cm]{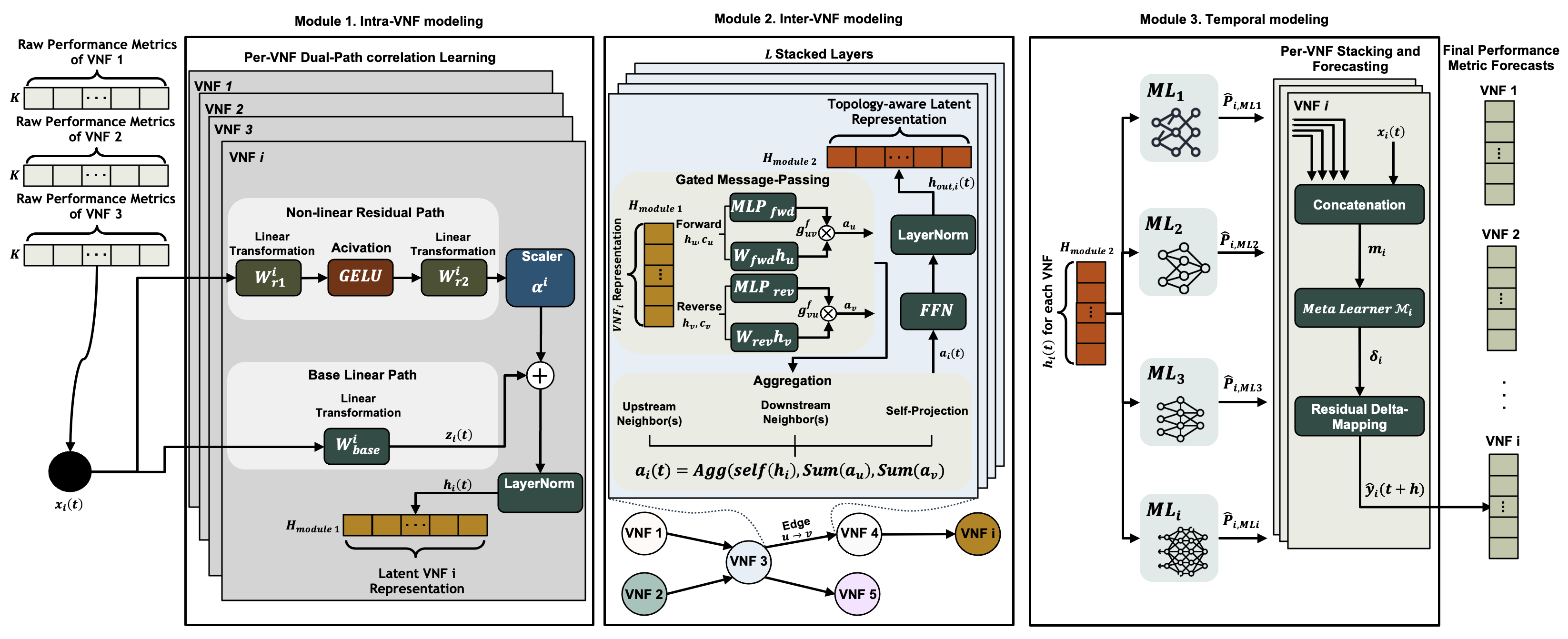}
    \caption{Architecture of \model, composed of three modules: Intra-VNF modeling, Inter-VNF modeling, and temporal modeling.}
    \label{fig:cefira_model}
\end{figure*}

\model, illustrated in Fig.~\ref{fig:cefira_model}, is organized as a sequential pipeline of three dedicated modules. Each module targets a distinct source of predictive information: Intra-VNF metric correlations, Inter-VNF topological dependencies, and heterogeneous temporal dynamics. The output of each module serves as the input to the next, and the entire pipeline is trained end-to-end by minimizing the mean absolute error between the final forecasts and the ground-truth performance metrics.

\subsection{System Model and Notation}

Let $G = (V, E)$ be a directed graph representing an SFC deployed in a cloud-native environment, where $V = \{v_1, \dots, v_N\}$ is the set of $N$ VNF nodes and $E$ encodes the directed traffic flow relationships between them.
At each discrete timestep $t$, every node $v_i \in V$ is associated with an input feature vector $\mathbf{x}_i(t) \in \mathbb{R}^{K}$, as defined in Eq.\eqref{eq:input_features}.
\begin{multline}
    \mathbf{x}_i(t) = \bigl[\,\text{CPU}_i(t),\;\text{memory}_i(t),\;
    \text{traffic}_i(t),\;c_i(t)\,\bigr] \\
    \in \mathbb{R}^{K}, \quad K = 4
    \label{eq:input_features}
\end{multline}
where $c_i(t) \in \{0,1\}$ is a binary crash indicator derived from the VNF status: $c_i(t) = 1$ if a crash is detected for VNF $v_i$ at timestep $t$, and $0$ otherwise. The crash indicator is included as an input feature only and is not itself a forecasting target.
The input to the forecasting model consists of a sliding window of the previous $T$ observations for each VNF, as defined in Eq.~\eqref{eq:input_window}.
\begin{equation}
    \mathbf{X}_i(t) = [\mathbf{x}_i(t-T+1), \dots, \mathbf{x}_i(t)] 
    \in \mathbb{R}^{T \times K}
    \label{eq:input_window}
\end{equation}
Using the historical observations from all VNFs together with the graph topology, the forecasting function forecasts the next-step performance metric values for each VNF according to Eq.~\eqref{eq:forecast}.
\begin{equation}
    \hat{\mathbf{y}}_i(t+h) = f_\theta\bigl(\{\mathbf{X}_j(t)\}_{j \in V}, G\bigr) 
    \in \mathbb{R}^{K^\prime}, \quad K^\prime = 3,
    \label{eq:forecast}
\end{equation}
where $K^\prime$ corresponds to the three target performance metrics: CPU usage, memory usage, and traffic load, and $h$ denotes the forecast horizon. 

\subsection{Module 1. Intra-VNF Modeling}

This module maps the raw input feature vector $\mathbf{x}_i(t)$ of each VNF $v_i$, defined in Eq.~\eqref{eq:input_features}, to a latent representation $\mathbf{h}_i(t) \in \mathbb{R}^H$, where $H$ is the hidden dimension used throughout \model. The mapping is defined in Eq.~\eqref{eq:feature_projection}.

\begin{equation}
    \mathbf{h}_i(t) = \text{LayerNorm}\Bigl( \mathbf{z}_i(t) + \alpha^{(i)} \cdot \mathcal{R}(\mathbf{x}_i(t)) \Bigr)
    \label{eq:feature_projection}
\end{equation}

The base linear projection (${z}_i(t)$) is defined in Eq.~\eqref{eq:base_projection}.
\begin{equation}
    \mathbf{z}_i(t) = \mathbf{W}_{\text{base}}^{(i)} \mathbf{x}_i(t) + \mathbf{b}_{\text{base}}^{(i)}
    \label{eq:base_projection}
\end{equation}

where $\mathbf{W}_{\text{base}}^{(i)} \in \mathbb{R}^{H \times K}$ is a learnable weight matrix that linearly projects the input performance metrics into the hidden space, and $\mathbf{b}_{\text{base}}^{(i)} \in \mathbb{R}^H$ is a learnable bias vector. This path captures linear correlations among the performance metrics.

The residual branch ($\mathcal{R}(\mathbf{x}_i(t))$) is defined in Eq.~\eqref{eq:residual_projection}.
\begin{equation}
    \mathcal{R}(\mathbf{x}_i(t)) = \mathbf{W}_{r2}^{(i)} \cdot \sigma\bigl(\mathbf{W}_{r1}^{(i)} \mathbf{x}_i(t) + \mathbf{b}_{r1}^{(i)}\bigr) + \mathbf{b}_{r2}^{(i)}
    \label{eq:residual_projection}
\end{equation}

where $\mathbf{W}_{r1}^{(i)} \in \mathbb{R}^{H \times K}$ and $\mathbf{W}_{r2}^{(i)} \in \mathbb{R}^{H \times H}$ are learnable weight matrices, $\mathbf{b}_{r1}^{(i)} \in \mathbb{R}^H$ and $\mathbf{b}_{r2}^{(i)} \in \mathbb{R}^H$ are learnable bias vectors, and $\sigma(\cdot)$ is the Gaussian Error Linear Unit (GELU) activation function, defined as $\sigma(z) = z \cdot \Phi(z)$, where $\Phi(z)$ is the standard Gaussian cumulative distribution function~\cite{hendrycks2016gaussian}. The activation introduces non-linearity, allowing the residual branch to capture non-linear correlations among the performance metrics that cannot be expressed by a purely linear mapping.

The residual scale $\alpha^{(i)} \in \mathbb{R}$ is a learnable scalar, that controls the contribution of the residual branch relative to the base projection. $\text{LayerNorm}(\cdot)$ denotes layer normalization, which normalizes the elements of its input vector to have zero mean and unit variance across the hidden dimension $H$. It is applied after the residual addition to stabilize the distribution of $\mathbf{h}_i(t)$ and improve training stability.

Each VNF $v_i$ is equipped with its own parameter set
$\Theta_i = \{ \mathbf{W}_{base}^{(i)}, \mathbf{b}_{base}^{(i)}, \mathbf{W}_{r1}^{(i)}, \mathbf{b}_{r1}^{(i)}, \mathbf{W}_{r2}^{(i)}, \mathbf{b}_{r2}^{(i)}, \alpha^{(i)} \}$. Because the parameters are not shared across VNFs, the module can learn distinct correlation patterns for different VNF types.

This design encodes the correlations among performance metrics directly into $\mathbf{h}_i(t)$ before any topological or temporal processing occurs. At the beginning of training, $\alpha^{(i)} \approx 0$, so $\mathbf{h}_i(t)$ is dominated by the linear path defined in Eq.~\eqref{eq:base_projection}. This provides a stable initialization and avoids large parameter updates that can occur when a non-linear function is trained from a random starting point. As training proceeds, $\alpha^{(i)}$ increases and the residual branch gradually contributes non-linear interactions. 

Algorithm~\ref{alg:intra-VNF} summarizes the computation of $\mathbf{h}_i(t)$ for a single timestep $t$ across all VNFs. For each VNF $v_i$, the algorithm first computes the base linear projection $\mathbf{z}_i(t)$ (Step 1), then evaluates the residual branch $\mathcal{R}_i(\mathbf{x}_i(t))$ via the two-layer network (Step 2), fuses the two paths using the residual scale $\alpha^{(i)}$ (Step 3), and finally applies layer normalization to obtain the latent representation $\mathbf{h}_i(t)$ (Step 4).

\begin{algorithm}
\caption{Intra-VNF Feature Projection}
\label{alg:intra-VNF}
    \textbf{Input:} Set of raw feature sequences for the SFC at timestep $t$, $\{ \mathbf{x}_i(t) \}_{i=1}^N$ \\
    \textbf{Output:} VNF-specific latent representation $\mathbf{h}_i(t) \in \mathbb{R}^H$
\begin{algorithmic}[1]    
    \State \textbf{for} {each VNF $v_i \in V$} \textbf{do}
        \State \quad Step 1: Primary Linear Projection
            \State \quad \quad $\mathbf{z}_i(t) \leftarrow \mathbf{W}_{base}^{(i)} \mathbf{x}_i(t) + \mathbf{b}_{base}^{(i)}$ 
        \State \quad Step 2: Non-linear Residual Processing
            \State \quad \quad $\mathbf{h}_{temp,i}(t) \leftarrow \mathbf{W}_{r1}^{(i)} \mathbf{x}_i(t) + \mathbf{b}_{r1}^{(i)}$ 
            \State \quad \quad $\mathbf{h}_{act,i}(t) \leftarrow \text{GELU}(\mathbf{h}_{temp,i}(t))$ 
            \State \quad \quad $\mathcal{R}_i(\mathbf{x}_i(t)) \leftarrow \mathbf{W}_{r2}^{(i)} \mathbf{h}_{act,i}(t) + \mathbf{b}_{r2}^{(i)}$ 
        \State \quad Step 3: Adaptive Feature Fusion
        \State \quad \quad $\mathbf{h}_{raw,i}(t) \leftarrow \mathbf{z}_i(t) + \alpha^{(i)} \cdot \mathcal{R}_i(\mathbf{x}_i(t))$
        \State \quad Step 4: Latent Normalization
            \State \quad \quad $\mathbf{h}_i(t) \leftarrow \text{LayerNorm}(\mathbf{h}_{raw,i}(t))$
    \State \textbf{end for}
    \State \Return $\{ \mathbf{h}_{i}(t) \}_{i=1}^N$
\end{algorithmic}
\end{algorithm}

\subsection{Module 2. Inter-VNF Modeling}

This module updates the latent representations $\mathbf{h}_i(t)$ produced by Module~1 by exchanging them between VNFs along the directed SFC graph $G = (V, E)$, using a Message-Passing Neural Network (MPNN)~\cite{gilmer2017neural}. It stacks $L$ layers with independent parameters. The equations below define one layer, which maps its input $\mathbf{h}_i(t)$ to an output $\mathbf{h}'_i(t)$ that is the input of the next layer. Since information relevant to forecasting can flow in both directions along an edge, each direction has its own gate and message weights.

For each directed edge $(u,v) \in E$, a forward gate scales the message from $u$ to $v$, and a reverse gate scales the message from $v$ to $u$, as defined in Eq.~\eqref{eq:gate_fwd} and Eq.~\eqref{eq:gate_rev}.
\begin{align}
    g_{uv}^f &= \text{Sigmoid}\bigl(\text{MLP}_{fwd}([\mathbf{h}_u(t); \mathbf{h}_v(t); c_u(t); c_v(t)])\bigr)
    \label{eq:gate_fwd} \\
    g_{vu}^r &= \text{Sigmoid}\bigl(\text{MLP}_{rev}([\mathbf{h}_v(t); \mathbf{h}_u(t); c_v(t); c_u(t)])\bigr)
    \label{eq:gate_rev}
\end{align}
where $[\cdot;\cdot;\cdot;\cdot]$ denotes vector concatenation, $c_u(t), c_v(t) \in \{0,1\}$ are the observed crash indicators of Eq.~\eqref{eq:input_features}, and $\text{Sigmoid}(z) = 1/(1+e^{-z})$ maps a scalar to $(0,1)$. $\text{MLP}_{fwd}(\mathbf{z}) = \mathbf{w}_{fwd}^{\top}\mathbf{z} + b_{fwd}$ and $\text{MLP}_{rev}(\mathbf{z}) = \mathbf{w}_{rev}^{\top}\mathbf{z} + b_{rev}$ are single linear layers with independent learnable weight vectors $\mathbf{w}_{fwd}, \mathbf{w}_{rev} \in \mathbb{R}^{2H+2}$ and biases $b_{fwd}, b_{rev} \in \mathbb{R}$. A gate value near $1$ lets a message pass in full, while a value near $0$ suppresses it.

Each VNF then aggregates its own state and the gated messages from its neighbors, as defined in Eq.~\eqref{eq:aggregation}.
\begin{multline}
    \mathbf{a}_i(t) = \mathbf{W}_{self}\mathbf{h}_i(t) + \mathbf{b}_{self}
    + \sum_{u:(u,i)\in E} g_{ui}^f \bigl(\mathbf{W}_{fwd}\mathbf{h}_u(t) + \mathbf{b}_{fwd}\bigr) \\
    + \sum_{v:(i,v)\in E} g_{vi}^r \bigl(\mathbf{W}_{rev}\mathbf{h}_v(t) + \mathbf{b}_{rev}\bigr)
    \label{eq:aggregation}
\end{multline}
where $\mathbf{a}_i(t) \in \mathbb{R}^H$ is the aggregated representation of VNF $v_i$, $\mathbf{W}_{self}, \mathbf{W}_{fwd}, \mathbf{W}_{rev} \in \mathbb{R}^{H \times H}$ are learnable weight matrices, and $\mathbf{b}_{self}, \mathbf{b}_{fwd}, \mathbf{b}_{rev} \in \mathbb{R}^H$ are learnable bias vectors. All gate and message parameters are shared across the edges of a layer. The first term preserves the VNF's own state, the first sum collects messages from its upstream neighbors, and the second sum collects messages from its downstream neighbors.

The aggregated representation is combined with the layer input through a residual connection and layer normalization ($\text{LayerNorm}$ as defined in Module~1), as defined in Eq.~\eqref{eq:spatial_fusion}.
\begin{equation}
    \mathbf{h}_{\text{spatial},i}(t) = \text{LayerNorm}\bigl(\mathbf{h}_i(t) + \mathbf{a}_i(t)\bigr)
    \label{eq:spatial_fusion}
\end{equation}
The result is refined by a point-wise feed-forward network (FFN), as defined in Eq.~\eqref{eq:ffn}.
\begin{equation}
    \text{FFN}(\mathbf{h}_{\text{spatial},i}(t)) = \mathbf{W}_{f2} \cdot 
\text{GELU}(\mathbf{W}_{f1}\mathbf{h}_{\text{spatial},i}(t) + 
\mathbf{b}_{f1}) + \mathbf{b}_{f2}
    \label{eq:ffn}
\end{equation}
where $\mathbf{W}_{f1} \in \mathbb{R}^{2H \times H}$ and $\mathbf{W}_{f2} \in \mathbb{R}^{H \times 2H}$ are learnable weight matrices, $\mathbf{b}_{f1} \in \mathbb{R}^{2H}$ and $\mathbf{b}_{f2} \in \mathbb{R}^H$ are learnable bias vectors, and $\text{GELU}(\cdot)$ is defined as in Module~1. The layer output is obtained through a second residual connection and layer normalization, as defined in Eq.~\eqref{eq:nonlinear_refinement}.
\begin{equation}
    \mathbf{h}'_i(t) = \text{LayerNorm}\bigl(\mathbf{h}_{\text{spatial},i}(t) + \text{FFN}(\mathbf{h}_{\text{spatial},i}(t))\bigr)
    \label{eq:nonlinear_refinement}
\end{equation}

The reverse message is included because a congested or crashed downstream VNF can affect the traffic that upstream VNFs are able to forward. Because the gates take the crash indicators as input, they can learn to reduce the influence of a neighbor that is currently crashing or recovering, limiting the propagation of corrupted signals through the SFC. Stacking $L$ layers lets the representation of a VNF be influenced by VNFs up to $L$ hops away.

Algorithm~\ref{alg:inter-vnf} summarizes the computation performed by a single layer: gate computation via Eqs.~\eqref{eq:gate_fwd}--\eqref{eq:gate_rev} (Step~1), gated aggregation via Eq.~\eqref{eq:aggregation} (Step~2), spatial feature fusion via Eq.~\eqref{eq:spatial_fusion} (Step~3), and non-linear latent refinement via Eqs.~\eqref{eq:ffn}--\eqref{eq:nonlinear_refinement} 
(Step~4).

\begin{algorithm}
\caption{Inter-VNF Modeling}
\label{alg:inter-vnf}
\textbf{Input:} Latent features $\{\mathbf{h}_i(t)\}_{i=1}^N$, SFC edge set $E$, crash indicators $\{c_i(t)\}_{i=1}^N$ \\
\textbf{Output:} Topology-aware latent representations $\{\mathbf{h}'_i(t)\}_{i=1}^N$
\begin{algorithmic}[1]
    \State Step 1: Gate Computation
    \State \textbf{for} each edge $(u, v) \in E$ \textbf{do}
    \State \quad $g_{uv}^f \leftarrow \text{Sigmoid}(\text{MLP}_{fwd}([\mathbf{h}_u(t); \mathbf{h}_v(t); c_u(t); c_v(t)]))$
    \State \quad $g_{vu}^r \leftarrow \text{Sigmoid}(\text{MLP}_{rev}([\mathbf{h}_v(t); \mathbf{h}_u(t); c_v(t); c_u(t)]))$
    \State \textbf{end for}
    \State Step 2: Gated Aggregation
    \State \textbf{for} each VNF $v_i \in V$ \textbf{do} 
    \State \quad compute $\mathbf{a}_i(t)$ via Eq.~\eqref{eq:aggregation} 
    \State \textbf{end for}
    \State Step 3: Spatial Feature Fusion
    \State \textbf{for} each VNF $v_i \in V$ \textbf{do} 
    \State \quad $\mathbf{h}_{spatial,i}(t) \leftarrow \text{LayerNorm}(\mathbf{h}_i(t) + \mathbf{a}_i(t))$ 
    \State \textbf{end for}
    \State Step 4: Non-linear Latent Refinement
    \State \textbf{for} each VNF $v_i \in V$ \textbf{do} 
    \State \quad $\mathbf{h}'_i(t) \leftarrow \text{LayerNorm}(\mathbf{h}_{spatial,i}(t) + \text{FFN}(\mathbf{h}_{spatial,i}(t)))$ 
    \State \textbf{end for}
    \State \Return $\{\mathbf{h}'_i(t)\}_{i=1}^N$
\end{algorithmic}
\end{algorithm}

\subsection{Module 3. Temporal Modeling}

This module maps the sequence of per-VNF latent representations produced by Module~2 to the final forecast of the three target performance metrics. 

The module applies four temporal backbones, $b \in \{\text{RNN}, \text{LSTM}, \text{GRU}, \text{Transformer}\}$, independently to $\mathbf{h}'_i(t)$, each followed by a linear output projection, to produce a backbone-specific forecast as defined in Eq.~\eqref{eq:backbone_forecast}.
\begin{equation}
    \mathbf{p}_{i,b}(t) = \mathbf{W}_{out}^{(b)}\,\text{Backbone}_b\bigl(\mathbf{h}'_i(t)\bigr) + \mathbf{b}_{out}^{(b)} \in \mathbb{R}^{K'}
    \label{eq:backbone_forecast}
\end{equation}
where $\text{Backbone}_b(\cdot)$ is the standard RNN, LSTM, GRU, or Transformer encoder~\cite{vaswani2017attention}, applied along the time dimension of $\mathbf{h}'_i(t)$, and $\mathbf{W}_{out}^{(b)} \in \mathbb{R}^{K' \times H}$, $\mathbf{b}_{out}^{(b)} \in \mathbb{R}^{K'}$ are the output projection's learnable weight matrix and bias vector for backbone $b$. We adopt a shared-weight design: $\text{Backbone}_b(\cdot)$, $\mathbf{W}_{out}^{(b)}$, and $\mathbf{b}_{out}^{(b)}$ are shared across all VNFs for a given $b$, rather than learned per-VNF. Because Modules~1 and~2 have already injected VNF-specific and topology-specific information into $\mathbf{h}_i(t)$, each VNF still presents a distinct input to the shared backbones, even though the backbone parameters themselves are not VNF-specific.

The four backbones are selected to capture complementary temporal dynamics. The standard RNN serves as a lightweight baseline for short-term fluctuations. LSTM and GRU are chosen for their gating mechanisms, which effectively capture long-range dependencies, while GRU offers a favourable complexity--accuracy trade-off. The Transformer encoder complements the recurrent models by using self-attention to directly model global patterns across the entire sequence~\cite{xiao2025st}. This diversity allows the ensemble to handle bursty, periodic, and non-stationary workload patterns more robustly than any single backbone.

To prevent the ensemble from collapsing onto a single dominant backbone, Gaussian noise is added to each backbone forecast during training, as defined in Eq.~\eqref{eq:meta_noise}.
\begin{equation}
\hat{\mathbf{p}}_{i,b}(t)=
\begin{cases}
\mathbf{p}_{i,b}(t)+\boldsymbol{\epsilon}_{i,b}(t),
& \boldsymbol{\epsilon}_{i,b}(t)\sim
\mathcal{N}(0,\sigma_{\text{meta}}^2)\quad \\\text{(training)}\\[2mm]
\mathbf{p}_{i,b}(t),
& \text{(inference)}
\end{cases}
\label{eq:meta_noise}
\end{equation}
where $\sigma_{\text{meta}} \in \mathbb{R}$ is a fixed hyperparameter (Table~\ref{tab:hyperparameters}) and $\boldsymbol{\epsilon}_{i,b}(t)$ is sampled independently for each VNF, backbone, and performance-metric dimension. 
Because this noise is added independently per backbone, the meta-learner cannot rely on any one backbone's forecast being noise-free, and is instead trained to combine the ensemble's consensus; the noise is disabled at inference ($\sigma_{\text{meta}} = 0$).

A dedicated meta-learner $\mathcal{M}_i$, implemented as a two-layer feed-forward network, is used for each VNF $v_i$ to combine the four backbone forecasts into the final prediction, as defined in Eq.~\eqref{eq:meta_learner}.
\begin{equation}
    \delta_i(t) = \mathbf{W}_2^{(i)} \,\text{ReLU}\bigl(\mathbf{W}_1^{(i)} \mathbf{m}_i(t) + \mathbf{b}_1^{(i)}\bigr) + \mathbf{b}_2^{(i)}
    \label{eq:meta_learner}
\end{equation}

where $\text{ReLU}(z)$ is the concatenation of the four backbone forecasts and the last observed input vector $\mathbf{x}_i(t)$ of Eq.~\eqref{eq:input_features}, $\mathbf{W}_1^{(i)} \in \mathbb{R}^{H_m \times (4K'+K)}$ and $\mathbf{W}_2^{(i)} \in \mathbb{R}^{K' \times H_m}$ are learnable weight matrices, $\mathbf{b}_1^{(i)} \in \mathbb{R}^{H_m}$ and $\mathbf{b}_2^{(i)} \in \mathbb{R}^{K'}$ are learnable bias vectors, and $H_m \in \mathbb{N}$ is the meta-learner's hidden dimension, a fixed hyperparameter (Table~\ref{tab:hyperparameters}). Because $\mathcal{M}_i$ has its own parameter set per VNF, it can specialize to the error characteristics of that VNF.

Rather than forecasting absolute values, $\mathcal{M}_i$ predicts a residual displacement $\delta_i(t)$ relative to the last observation, as defined in Eq.~\eqref{eq:delta_mapping}.
\begin{equation}
    \hat{\mathbf{y}}_i(t+h) = \mathbf{x}_i(t)[1{:}K'] + \delta_i(t)
    \label{eq:delta_mapping}
\end{equation}
where $\mathbf{x}_i(t)[1{:}K']$ denotes the first $K'$ entries of $\mathbf{x}_i(t)$, i.e., the CPU, memory, and traffic values of Eq.~\eqref{eq:input_features}, excluding the crash indicator $c_i(t)$, which is not a forecast target. This residual formulation reduces the impact of non-stationarity: rather than learning the absolute scale of each metric from scratch, $\mathcal{M}_i$ only needs to learn the change from the last observed value.

Algorithm~\ref{alg:temporal_meta} summarizes the procedure. In Step~1, each backbone produces an initial forecast. In Step~2 (training only), Meta-Input Noise is added. In Step~3, the per-VNF meta-learner then computes the residual displacement, which is added to the last observed performance-metric values to obtain the final forecast.

\begin{algorithm}
\caption{Heterogeneous Temporal Ensemble (Module 3)}
\label{alg:temporal_meta}
\textbf{Input:} Per-VNF latent sequences $\{\mathbf{h}'_i(t)\}_{i=1}^N$; last observed inputs $\{\mathbf{x}_i(t)\}_{i=1}^N$ \\
\textbf{Output:} Forecasts $\{\hat{\mathbf{y}}_i(t+h)\}_{i=1}^N$
\begin{algorithmic}[1]
    \State Step 1: Per-Backbone Forecasting
    \State \textbf{for} each VNF $v_i \in V$ \textbf{do}
        \State \quad \textbf{for} each backbone $b$ \textbf{do}
            \State \quad \quad $\mathbf{p}_{i,b}(t) \leftarrow \mathbf{W}_{out}^{(b)}\,\text{Backbone}_b(\mathbf{h}'_i(t)) + \mathbf{b}_{out}^{(b)}$
        \State \quad \textbf{end for}
    \State \textbf{end for}

    \State Step 2: Meta-Input Noise
    \State \textbf{if} training \textbf{then}
        \State \quad $\hat{\mathbf{p}}_{i,b}(t) \leftarrow \mathbf{p}_{i,b}(t) + \boldsymbol{\epsilon}_{i,b}(t)$, $\boldsymbol{\epsilon}_{i,b}(t) \sim \mathcal{N}(0,\sigma^2_{\text{meta}} )$
    \State \textbf{else}
        \State \quad $\hat{\mathbf{p}}_{i,b}(t) \leftarrow \mathbf{p}_{i,b}(t)$
    \State \textbf{end if}

    \State Step 3: Per-VNF Stacking and Delta Forecasting
    \State \textbf{for} each VNF $v_i \in V$ \textbf{do}
        \State \quad $\mathbf{m}_i(t) \leftarrow \text{Concat}(\hat{\mathbf{p}}_{i,\text{RNN}}(t), \hat{\mathbf{p}}_{i,\text{LSTM}}(t), \hat{\mathbf{p}}_{i,\text{GRU}}(t),$ 
        \Statex \hspace{8em} $\hat{\mathbf{p}}_{i,\text{Transformer}}(t), \mathbf{x}_i(t))$
        \State \quad $\delta_i(t) \leftarrow \mathcal{M}_i(\mathbf{m}_i(t))$
        \State \quad $\hat{\mathbf{y}}_i(t+h) \leftarrow \mathbf{x}_i(t)[1{:}K'] + \delta_i(t)$
    \State \textbf{end for}
    \State \Return $\{\hat{\mathbf{y}}_i(t+h)\}_{i=1}^N$
\end{algorithmic}
\end{algorithm}

\section{Experimental Setup}
\label{sec:experimental_setup}

Evaluating \model requires a testbed that generates real Kubernetes-triggered crash events, instantiates the directed SFC graph with real inter-VNF traffic dependencies, and produces the four input features of Eq.~\eqref{eq:input_features} at scale and under varied operating conditions. 
This section describes the experimental testbed, which instantiates the general system architecture of Fig.~\ref{fig: High Level Architecture} using an IoT-based workload instantiation and a cloud-native NFV deployment. The testbed includes workload generation, traffic delivery to the VNFs, and the VNF Performance Forecasting Framework responsible for monitoring and forecasting VNF resource usage.
The testbed implementation, including the VNF containers, Kubernetes manifests, IoT traffic generator, and performance monitoring service, together with the collected VNF performance dataset, is publicly available.\footnote{\url{https://github.com/mld-rafiee/vnf-sfc-monitoring}}

\subsection{IoT-based Workload Generation}
\label{sec:data_gen_layer}

In the experimental instantiation of the Traffic and Workload Generation layer, software-emulated IoT devices and gateways are used to generate real workload streams for the SFC. The IoT devices periodically generate sensor messages at configurable rates, while the gateways aggregate these messages into batches before forwarding the resulting traffic to the computing continuum. This design follows the edge-based real-time IoT analytics testbed of Tusa et al.~\cite{tusa2024microservices}, where software IoT devices generate data streams and software IoT gateways operate as aggregation points. The software-based emulation provides scalability and reproducibility without requiring large-scale physical IoT deployments.

The resulting traffic determines traffic load, and consequently the CPU and memory usage that \model is trained to forecast. To evaluate \model under a range of realistic operating conditions, we designed a set of experiments emulating four representative IoT deployment scales, small, medium, large, and very large, constructed by varying the number of IoT devices and gateways, as summarized in Table~\ref{tab:exp}.

The small scenario corresponds to a residential smart-home setting, with one gateway serving 1,000 devices, consistent with smart-home gateway architectures that allow end devices to be added dynamically on the gateway~\cite{chattopadhyay2018device}. The medium scenario represents commercial or enterprise environments such as office buildings or retail spaces, with 10 gateways and 10,000 devices, aligned with scalable IoT control in commercial-building HVAC systems~\cite{png2019internet}. The large scenario emulates industrial IoT deployments, with 25 gateways and 25,000 devices, and is supported by gateway-based IIoT work targeting large-scale, multi-device deployments~\cite{ding2025semi}. Finally, the very large scenario emulates city-scale smart-city infrastructure, with 50 gateways and 50,000 devices, reflecting AI- and IoT-driven city-scale sensing work for roadside infrastructure maintenance~\cite{forkan2024aiotcitysense}.

\begin{table}[ht]
    \fontsize{7.3pt}{9pt}\selectfont
    \centering
    \caption{Experimental configurations for IoT traffic generation}
    \label{tab:exp}
    \begin{tabular}{@{}l c c c c}
        \hline
        \multirow{2}{*}{\textbf{Parameter}} & \multicolumn{4}{c}{\textbf{Scenario}} \\ \cline{2-5} 
        & \textbf{Small} & \textbf{Medium} & \textbf{Large} & \textbf{Very Large}  \\ \hline
        IoT Gateway & 1 & 10 & 25 & 50 \\ 
        IoT Devices & 1000 & 10000 & 25000 & 50000 \\ \hline
        Sensor Rate (messages/s) & \multicolumn{4}{c}{4, 12, 25, 50} \\ 
        Data batch size & \multicolumn{4}{c}{4000} \\ \hline
    \end{tabular}
    \vspace{-5pt}
\end{table}

IoT sensor data streams are generated as JSON-formatted messages containing a sensor identifier, name, type, measurement unit, value, and timestamp. Sensors produce messages at rates ranging from 4 to 50 per second, and gateways aggregate them into batches of size $b_s = 4000$. This range enables the generation of progressively larger volumes of sensor data, exercising \model under increasingly demanding and varied traffic conditions.

\vspace{-5pt}
\subsection{RAN / Mobile Core}
\label{sec:ran_mobile_core}

As introduced in Section~\ref{sec:high_level_architecture}, IoT traffic traverses the RAN/Mobile Core before reaching the computing continuum, and the resulting traffic load is one of the four input features of \model. Rather than relying on synthetic bandwidth assumptions, we drive this segment of the testbed with empirical 4G/5G measurements, consistent with the growing use of cellular connectivity for IoT deployments~\cite{moges2023cellular}, so that the traffic-load input, and consequently the CPU and memory usage \model is trained to forecast, reflects realistic variability and constraints observed in operational 4G/5G networks. Since IoT traffic in our scenario flows from devices toward the processing node, we configure the traffic rate delivered to the mobile network using uplink data-rate statistics collected by the NorNet Edge (NNE) measurement platform, deployed across Norway~\cite{ocampo2024reinforcement}.
The NNE dataset was collected between November 2023 and February 2024 and reports bandwidth-related performance metrics, including downlink and uplink data rates from aggregated speed tests across both 4G and 5G networks.

\vspace{-5pt}
\subsection{Computing Continuum}
\label{sec:computing_continuum}

As shown in Fig.~\ref{fig: High Level Architecture}, IoT traffic reaching the Computing Continuum is processed by an ordered chain of VNFs before reaching its destination. IoT traffic commonly traverses security and processing functions such as firewalls, DPI, and NAT before reaching the target service~\cite{xu2020qos, jia2025efficient}. We instantiate this chain using five representative functions, Firewall, 
DPI, Encryption, Compression, and NAT, arranged into two SFC topologies under static and dynamic computing resource configurations.

\subsubsection{SFC Topologies}
\label{sec:sfc_topology}
We evaluate \model under two SFC topologies, a linear four-node non-shared topology and a six-node shared-VNF topology. This selection targets our central hypothesis that Inter-VNF topological dependency is necessary for accurate forecasting: the non-shared topology provides a single linear dependency path, while the shared-VNF topology introduces resource contention and traffic aggregation at a shared VNF instance, allowing the crash-aware gating of Module~2 to be tested under conditions where one VNF's failure affects two downstream SFCs at once.

\begin{figure}[htbp]
    \centering
    \includegraphics[scale=0.45]{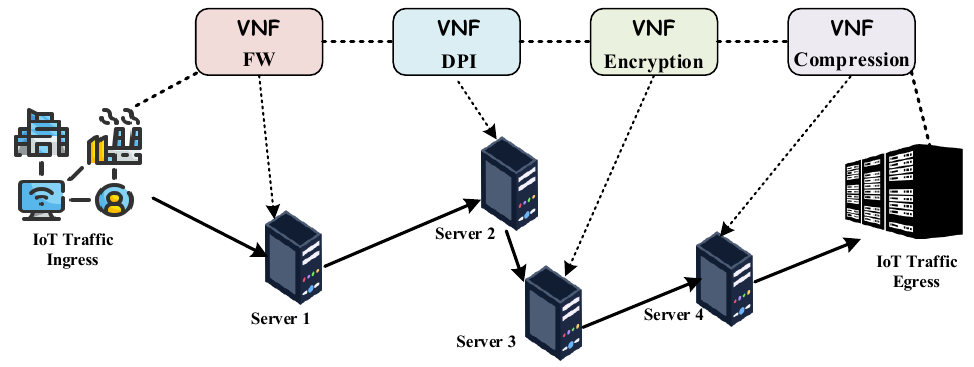}
    \caption{Non-shared VNF: a linear four-node SFC, where each VNF instance serves a single SFC.}
    \label{fig: non-shared vnf}
\end{figure}

\paragraph{Non-shared VNF}
The first topology is a linear four-node SFC ($N=4$), shown in Fig.~\ref{fig: non-shared vnf}, in which each VNF instance is dedicated to a single SFC. This topology is a security-oriented SFC combining traffic filtering, inspection, and confidentiality/efficiency processing, of the kind adopted in security-aware NFV deployments~\cite{Wang2024SecureSFC}.
The Firewall performs coarse-grained stateful packet filtering at Layers~3--4, dropping malformed or policy-violating traffic before it reaches computationally expensive downstream functions~\cite{Halpern2015RFC7665}. DPI follows immediately to perform application-layer inspection on clean plaintext: this ordering is a hard constraint, since IDS/DPI must inspect payloads before compression or encryption, as ciphertext is opaque to any inspection engine~\cite{Xu2019Availability}. Encryption is then applied to guarantee confidentiality in transit; because intermediate VNFs require access to the payload they process, SFC traffic cannot be end-to-end encrypted from the source, and encryption must instead be applied at a defined point in the SFC, after inspection is complete~\cite{Gunleifsen2017SFCEncryption}. Compression is placed last to reduce outgoing link utilization, applied to transport-header and encapsulation overhead rather than the encrypted payload itself, which resists further lossless compression.

\begin{figure}[htbp]
    \centering
    \includegraphics[scale=0.45]{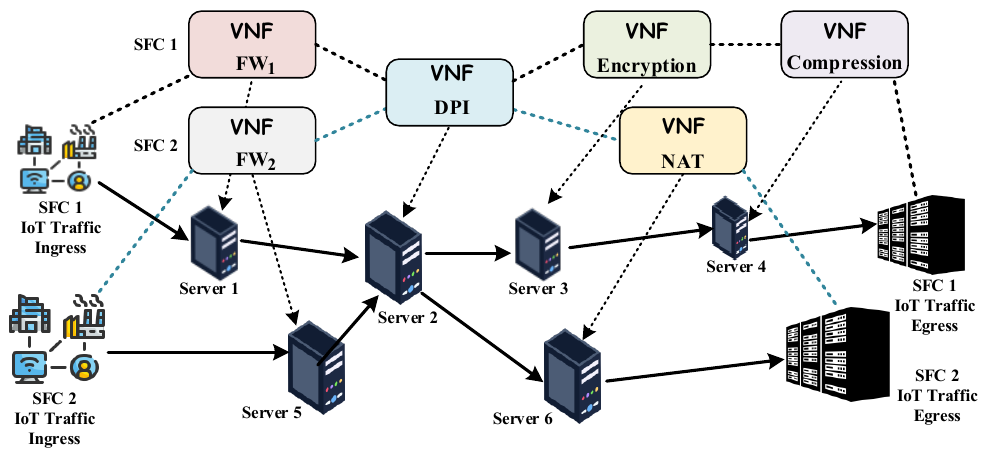}
    \caption{Shared-VNF SFC: a six-node SFC in which two SFCs share a common DPI instance.}
    \label{fig: shared vnf}
\vspace{-5pt}
\end{figure}

\paragraph{Shared-VNF}
The second topology introduces a six-node SFC ($N=6$) with a shared prefix, shown in Fig.~\ref{fig: shared vnf}. Two SFCs share a common DPI instance placed after their respective firewalls. 
$\text{FW}_1$ and $\text{FW}_2$ enforce distinct policies for their respective traffic classes, while the single DPI instance serves both chains. 
The first SFC applies Encryption followed by Compression after DPI inspection, targeting security- and bandwidth-sensitive traffic; the second applies NAT after DPI inspection, targeting inter-domain routing traffic that requires address translation. Shared-VNF configurations of this type are a key motivation for NFV, reducing CAPEX and OPEX by reusing function instances across SFCs while preserving ordering constraints~\cite{Mijumbi2016NFV, mijumbi2017topology, mohamad2022edge, malandrino2019reducing, guo2020cost, wang2025vnf, abdulghaffar2024impact, yi2017generalized, mohamad2020psvshare}.

\subsubsection{VNF Implementation}
\label{sec:vnf_implementation}
The experimental testbed consists of a cluster of four to six nodes, each node hosts one containerized VNF, running inside a container that packages its software together with all required dependencies through Docker~\cite{Docker}. Kubernetes~\cite{Kubernetesframework} additionally provides the traffic routing and service discovery that steer packets through each VNF of the SFC in sequence.

The Firewall relies on a JSON-based configuration file to set blocking rules based on the packet's 5-tuple, namely source IP, destination IP, source port, destination port, and protocol. The DPI, built on Snort3 (version 3.12.2.0)~\cite{Roesch1999Snort}, inspects the payloads of packets passing through the network. The Encryption VNF relies on Advanced Encryption Standard -- Galois/Counter Mode (AES-GCM) to protect network traffic~\cite{Dworkin2007GCM}, providing both encryption and authentication so that packet confidentiality and integrity are maintained. The Compression VNF compresses the packet payload using the Brotli algorithm~\cite{Alakuijala2016Brotli}, followed by base64 encoding for protocol-safe transmission. The NAT carries out source NAT only on traffic coming from a designated ingress IP address, translating the source IP into a public NAT address, and dynamically mapping the original source ports to an available port drawn from a configurable port pool.

\subsubsection{Computing Resource Configurations}
\label{sec:resource_alloc}

The CPU and memory limits assigned to each VNF by Kubernetes, which we refer to collectively as computing resources, directly shape the resource-usage traces that \model is trained to forecast, and, as discussed in Section~\ref{sec:high_level_architecture}, exceeding these limits is what triggers throttling or an OOMKilled restart. To evaluate \model under both a stationary and a non-stationary resource limit, we consider two configurations that differ in whether these limits remain fixed or change over the course of an experiment: \emph{Static} and \emph{Dynamic}.

\paragraph{Static Computing Resource Configuration}
In this configuration, CPU and memory limits are fixed for each VNF for the entire duration of the experiment. Because the limit itself never changes, throttling and OOMKilled events under this configuration are driven purely by rising traffic-driven demand against a fixed limit. These events are the primary source of the crash events captured by the crash status metric introduced in Section~\ref{sec:high_level_architecture} and used as the crash indicator $c_i(t)$ in \model (Eq.~\eqref{eq:input_features}).

\paragraph{Dynamic Computing Resource Configuration}
In this configuration, the CPU and memory limits of each VNF are instead adjusted over time, while traffic intensity is simultaneously varied. This emulates operator- or orchestrator-driven resource changes and is intended to test \model's ability to forecast resource demand when the resource limit itself, and not only the workload, is time-varying.

Resource adjustments are sampled from a mixture of four vertical scaling profiles: steady-state, ramp, step, and spike. steady-state periods hold the limit constant. The ramp profile gradually adjusts the limit over multiple steps, representing planned or controlled scaling. The step profile applies an abrupt change to the limit, representing operator interventions or orchestration decisions. The spike profile applies a temporary boost, mainly to the CPU limit with a smaller increase to the memory limit, representing a rapid response to a sudden traffic increase. Together, these four profiles are designed to cover representative vertical scaling behaviors in cloud and NFV environments: prolonged stability, gradual adjustment, sudden change, and short-term boost~\cite{rankothge2019scaling}.

\subsection{VNF Performance Forecasting Framework}
\label{sec:vnf_forecasting_framework}

As introduced in Section~\ref{sec:high_level_architecture}, the VNF Performance Forecasting Framework resides in the management and orchestration plane and consists of the Performance Monitoring Component and the Forecasting Component (\model). Section~\ref{sec:monitoring_implementation} describes the implementation of the former, and Section~\ref{sec:model_implementation} the configuration of the latter.

\subsubsection{Performance Monitoring Implementation}
\label{sec:monitoring_implementation}
This subsection describes the implementation of the Performance Monitoring Component and its constituent monitoring modules, which monitor the four input features of \model for each VNF. Because these metrics originate from heterogeneous sources, Kubernetes-based monitoring and API-based reporting are combined into a hybrid monitoring strategy.

Prometheus~\cite{prometheus}, a widely adopted open-source monitoring tool, queries the Kubernetes monitoring infrastructure at regular intervals to retrieve CPU usage, memory usage, and crash status via the Kubernetes API for each VNF; traffic load, self-reported by each VNF container, is instead monitored through a dedicated REST API. This split allows the framework to rely on standardized cluster monitoring for resource telemetry while supporting an application-specific metric, traffic load, that is not natively exposed by Kubernetes.
For traffic load, a REST API provides a standardized interface for VNFs to report their metrics via HTTP. The traffic load is computed in Mbps from the number of bytes processed during the reporting interval and sent as a JSON payload to a predefined REST endpoint. This design keeps monitoring overhead low and allows each VNF to report independently.

Performance metrics are sampled at a temporal granularity of 30 seconds, with each observation corresponding to aggregated telemetry over a 30-second interval for each VNF in both topologies. This granularity preserves short-term dynamics relevant to resource contention and workload bursts, while ensuring stable and tractable time-series inputs for machine learning models. It is also consistent with the native sampling interval of the widely adopted Clearwater vIMS dataset~\cite{clearwater2018}, the de facto dataset for VNF resource prediction and scaling studies~\cite{mijumbi2017topology,xiao2025st}.
The resulting dataset is partitioned into 80\% for training, and 20\% for testing.

\subsubsection{\model Implementation}
\label{sec:model_implementation}

This subsection specifies the hyperparameter configuration of the \model, used throughout the experiments of Section~\ref{sec:evaluationresult}. To ensure a consistent experimental setting across all topologies and computing resource configurations, the same hyperparameter configuration, summarized in Table~\ref{tab:hyperparameters}, is used in every experiment.

\begin{table}[htbp]
\centering
\caption{\model hyperparameter configuration.}
\label{tab:hyperparameters}
\fontsize{7.3pt}{7.5pt}\selectfont
\begin{tabular}{llc}
\toprule
\textbf{Component} & \textbf{Parameter} & \textbf{Value} \\
\midrule
\multirow{3}{*}{Global}
  & Sliding Window ($T$)                  & 30 \\
  & Hidden Dimension ($H$)                & 256 \\
  & Forecast Horizon ($h$)                & 30--120\,s \\
\midrule
\multirow{2}{*}{Module 1}
  & Residual Scale $\alpha^{(i)}$ init    & 0.01 \\
  & Activation                            & GELU \\
\midrule
Module 2
  & GNN Layers ($L$)                      & 2 \\
\midrule
\multirow{7}{*}{Module 3}
  & \multirow{2}{*}{Backbones}  & RNN, LSTM, \\
  &                             & GRU, Transformer \\
  & Backbone Layers                       & 1 \\
  & Transformer Attention Heads           & 4 \\
  & Meta-learner Hidden Dim ($H_m$)       & 16 \\
  & Meta-learner Activation               & ReLU \\
  & Meta-noise ($\sigma_{\text{meta}}$)   & 0.02 \\
\midrule
\multirow{6}{*}{Training}
  & Optimizer                             & Adam \\
  & Weight Decay                          & $10^{-5}$ \\
  & Batch Size                            & 128 \\
  & Early Stopping Patience               & 50 \\
  & Loss Function                         & MAE \\
  & Learning Rate ($\eta$)               & $10^{-3}$ \\
\bottomrule
\end{tabular}
\vspace{-10pt}
\end{table}

\section{Evaluation Results}
\label{sec:evaluationresult}

This section evaluates \model along three lines: the contribution of each module and the effect of crash-aware gating on forecasting accuracy; accuracy relative to existing topology-aware baselines; and inference latency and multi-horizon accuracy as indicators of deployability for proactive orchestration. All results are reported across the four topology--configuration combinations of 
Section~\ref{sec:experimental_setup}, using MAE, RMSE, and $R^2$ computed jointly over all VNFs and target metrics.

\subsection{Architectural Contribution and Crash-Awareness}
\label{sec:eval_architecture}

This subsection examines the source of \model's accuracy: the contribution of each of its three modules, the target metric that benefits most from each, and the specific contribution of the crash indicator within the Inter-VNF module.

\subsubsection{Contribution of Individual Modules}
\label{sec:eval_three_modules}

We build \model up one module at a time: the Temporal Ensemble (Module~3) alone, which has no information about a VNF's own metric correlations or about neighboring VNFs; adding the Intra-VNF module (Module~1), which encodes a VNF's own metric correlations but still ignores neighboring VNFs; and adding the Inter-VNF module (Module~2), which propagates state across the SFC graph and suppresses signals from crashing neighbors. Table~\ref{tab:eval_three_modules} reports MAE, RMSE, and $R^2$ for these four configurations across all four topology--configuration combinations.

Module~3 alone already reaches $R^2$ between 0.8899 and 0.9505, using only the diversity of the four backbones and the residual delta-mapping of the meta-learner (Eq.~\eqref{eq:delta_mapping}), with no explicit topology or metric-correlation information. This follows from the complementary inductive biases of the four backbones (Section~\ref{sec:forecasting_model}): the recurrent models capture short-term fluctuations and longer-range dependencies, while the Transformer's self-attention captures global patterns across the sequence. This stacking step already reduces the error any one backbone would make on its own, which is why a bare temporal ensemble is a reasonably strong baseline even before Modules~1 and~2 add topology and metric-correlation information.

Adding Module~1 reduces MAE consistently, by 4.0--13.9\% depending on the combination, validating the dual-path residual encoder's ability to capture cross-metric couplings, such as the leading influence of traffic load on subsequent CPU and memory utilization, prior to temporal processing; the gain is smallest under shared-VNF/dynamic, reflecting the difficulty of resolving a VNF's own metric correlations when the resource limit itself is also changing.

Adding Module~2 instead of Module~1 produces the largest single-module reduction observed, 16.8--24.8\% depending on the combination, consistently larger than Module~1. This asymmetry holds in every combination and grows more pronounced in Shared-VNF and Dynamic settings, indicating that Inter-VNF dependencies and failure propagation carry more predictive information for SFC forecasting than a VNF's own intra-node correlations: a temporal model can partially infer intra-VNF relationships from its own history, but has no access to the topological dependencies encoded in graph $G$ unless Module~2 provides them. The \model, with all three modules, achieves the best $R^2$ in every combination (0.9421--0.9808), confirming the three modules are complementary rather than redundant.

\begin{table*}[htbp]
\centering
\caption{Average performance of \model across topologies and resource allocation configurations, by module composition}
\label{tab:eval_three_modules}
{\fontsize{10}{12}\selectfont
\resizebox{\textwidth}{!}{%
\begin{tabular}{l  ccc ccc ccc ccc }
\toprule
Modules & \multicolumn{3}{c}{Non-shared VNF/Static} & \multicolumn{3}{c}{Non-shared VNF/Dynamic} & \multicolumn{3}{c}{Shared-VNF/Static} & \multicolumn{3}{c}{Shared-VNF/Dynamic}  \\
\cmidrule(lr){2-4} \cmidrule(lr){5-7} \cmidrule(lr){8-10} \cmidrule(lr){11-13} 
 & MAE & RMSE & $R^2$ & MAE & RMSE & $R^2$ & MAE & RMSE & $R^2$ & MAE & RMSE & $R^2$  \\
\midrule

\model (Module 3)              
                         & 0.038505 & 0.085719 & 0.889923
                         &0.015906 & 0.058654 & 0.950525   
                         &0.031084 & 0.075196 & 0.920956 
                         &0.038881 & 0.112698 & 0.909293 \\
                        
\model (Module 1 + Module 3) 
                        &                       
                            0.033135 & 0.072497 & 0.921263 
                          &  0.014101 & 0.050633 & 0.964158
                          &  0.027951 & 0.069647 & 0.934700
                          &  0.037344   &  0.111156    & 0.911757\\

\model (Module 2 + Module 3) &  
                         
                        0.030201 & 0.066221 & 0.934304 
                        &0.013239   & 0.048745 & 0.973251
                        &0.023375 & 0.056400 & 0.955533
                        & 0.031715 & 0.094056 & 0.936819\\

\model (All Modules)           & 
                        
                        0.029143 & 0.059195 & 0.947505 
                        &0.012275 & 0.041925  & 0.980775
                        &0.022223 & 0.049290 & 0.966037
                        &0.030833  & 0.090051 & 0.942085  \\
\bottomrule
\end{tabular}
}}
\vspace{-10pt}
\end{table*}

\subsubsection{Per-Metric Forecasting Performance}
\label{sec:per_performance_metric}

\begin{figure}[htbp]
    \centering
    \includegraphics[scale=0.48]{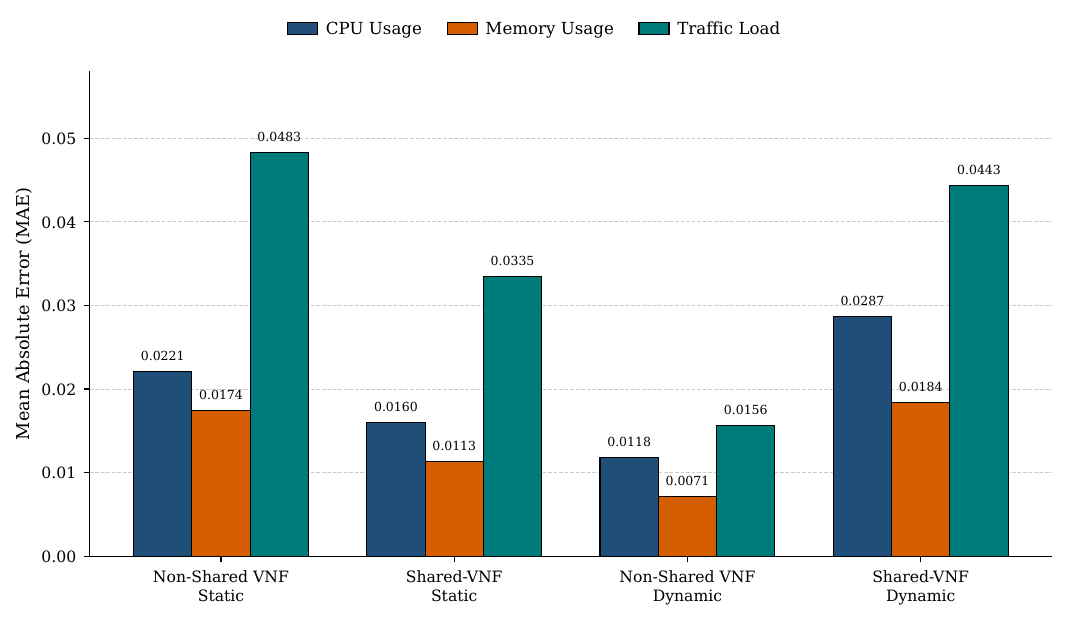}
    \caption{Forecasting performance of \model, broken down by target performance metric, across topologies and resource allocation configurations.}
    \label{fig:per_performance_metric_result}
\end{figure}

Fig.~\ref{fig:per_performance_metric_result} reports MAE separately for CPU usage, memory usage, and traffic load, to identify which metric benefits most from the Inter-VNF module.

Across all four combinations, memory usage is forecast with the lowest error, followed by CPU usage, with traffic load consistently the hardest. In Non-shared/Dynamic, \model reaches an MAE of 0.0118 for CPU, 0.0071 for memory, and 0.0156 for traffic load; error rises for all three in Shared-VNF/Dynamic, most visibly for traffic load and memory (0.0443 and 0.0184). This is a property of the scenario, not a weakness of \model: at a shared VNF such as the common DPI instance, traffic load reflects the merged demand of two SFCs rather than one predictable upstream source, and under the Dynamic configuration, memory usage additionally reflects crash-and-recovery transients as the resource limit changes over time. This confirms that traffic load, the most topology-dependent of the three metrics, is what Module~2's dominance in Section~\ref{sec:eval_three_modules} is largely driven by, and that memory's sensitivity to Shared-VNF/Dynamic specifically motivates the crash-indicator analysis below.

\subsubsection{Effect of the Crash Indicator}
\label{sec:crash_impact}

This subsection isolates the contribution of crash-aware gating by comparing \model against a variant with the crash indicator $c_i(t)$ removed from the input feature vector, leaving the rest of the architecture unchanged. We report this comparison under the Dynamic configuration, which produces a higher and more varied crash rate than Static and therefore gives a more informative test of the indicator's effect. Table~\ref{tab:with_without_crash_comparison} reports MAE, RMSE, and $R^2$ per VNF, with and without the indicator; a lower MAE/RMSE or higher $R^2$ in the "with" columns indicates the indicator is helping.

\begin{figure}[htbp]
\vspace{-5pt}
    \centering
    \includegraphics[scale=0.5]{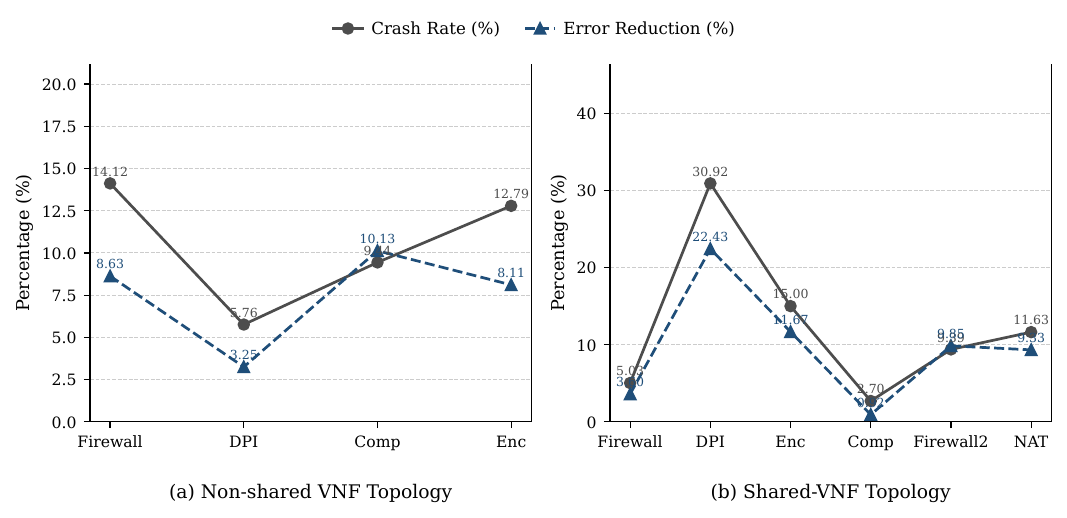}
    \caption{Per-VNF crash rate versus MAE reduction from the crash indicator: (a) Non-shared VNF topology and (b) Shared-VNF topology.}
    \label{fig:crash_percentage}
    \vspace{-10pt}
\end{figure}

The crash indicator improves accuracy at every VNF, in both topologies: average MAE improves by $\sim$7.5\% in the Non-shared topology and $\sim$12.7\% in the Shared-VNF topology, the larger gain matching Section~\ref{sec:eval_three_modules}'s finding that Inter-VNF coupling matters more under Shared-VNF. Fig.~\ref{fig:crash_percentage} plots each VNF's observed crash rate against its MAE reduction; the benefit tracks how often a VNF actually crashes rather than being uniform. The clearest case is the shared DPI instance (Section~\ref{sec:sfc_topology}): at a 30.9\% crash rate, MAE drops by 22.4\% (0.0758 to 0.0588) once the indicator is added. Because this VNF serves two SFCs at once, a crash there removes a reliable signal from both chains simultaneously, precisely the failure mode the crash-aware gating of Module~2 is designed to suppress; the result confirms it does so most at the VNF where this failure mode is most frequent, directly supporting the paper's central crash-ambiguity motivation (Section~\ref{sec:introduction}): without $c_i(t)$, the model cannot distinguish low load from a crash-and-recovery transient.

\begin{table}[htbp]
\caption{Per-VNF forecasting performance with and without the crash indicator.}
\label{tab:with_without_crash_comparison}
\centering
\resizebox{\columnwidth}{!}{
\begin{tabular}{l l lll lll lll}
\toprule
& & \multicolumn{3}{c}{\model without crash indicator} & \multicolumn{3}{c}{\model with crash indicator} \\
\cmidrule(lr){3-5} \cmidrule(lr){6-8}
Topology & VNF & {MAE} & {RMSE} & {R²} & {MAE} & {RMSE} & {R²} \\
\midrule
\multirow{5}{*}{Non-shared VNF} 
                            & Firewall      & {0.0255} & {0.0803} & {0.9450} & {0.0233} & {0.0791} & {0.9550} \\
                            & DPI           & {0.0123} & {0.0373} & {0.9812} & {0.0119} & {0.0356} & {0.9844} \\
                            & Enc           & {0.0074} & {0.0238} & {0.9910} & {0.0068} & {0.0233} & {0.9932} \\
                            & Comp          & {0.0079} & {0.0305} & {0.9882} & {0.0071} & {0.0297} & {0.9905} \\
                            
\midrule
\multirow{7}{*}{Shared-VNF} 
                            & Firewall  & {0.0278} & {0.0768} & {0.8047} & {0.0268} & {0.0712} & {0.8320} \\
                            & DPI       & {0.0758} & {0.1816} & {0.7709} & {0.0588} & {0.1397} & {0.8645} \\
                            & Enc       & {0.0240} & {0.0748} & {0.9621} & {0.0212} & {0.0688} & {0.9679} \\
                            & Comp      & {0.0218} & {0.0618} & {0.9720} & {0.0216} & {0.0617} & {0.9721} \\
                            & Firewall2 & {0.0325} & {0.0958} & {0.9307} & {0.0293} & {0.0854} & {0.9449} \\
                            & NAT       & {0.0300} & {0.0978} & {0.9372} & {0.0272} & {0.0908} & {0.9459} \\
                          
\bottomrule
\end{tabular}
}
\vspace{-10pt}
\end{table}

\subsection{Comparison with State-of-the-Art Methods}
\label{sec:compare_model}

\begin{figure}[htbp]
\vspace{-7pt}
    \centering
    \includegraphics[scale=0.5]{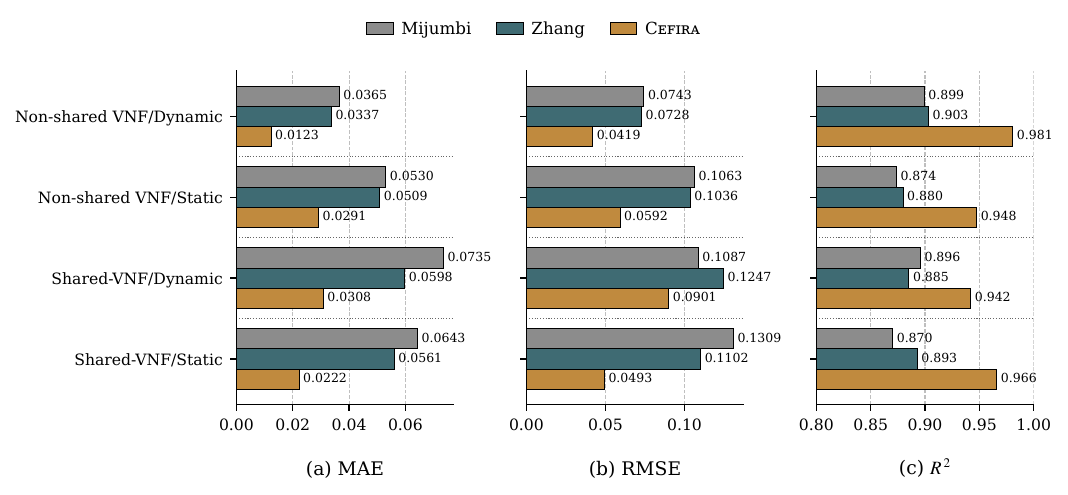}
    \caption{\model versus Mijumbi ~\cite{mijumbi2017topology} and Zhang ~\cite{zhang2023forecast} across both topologies and resource configurations: (a) MAE, (b) RMSE, and (c) $R^2$.}
    \label{fig:compare_model}
\end{figure}

We compare \model against two topology-aware baselines, Mijumbi et al.~\cite{mijumbi2017topology} and Zhang et al.~\cite{zhang2023forecast}, 
both of which explicitly model SFC topology and jointly forecast multiple performance metrics. Unlike \model, neither incorporates Intra-VNF metric correlations or a crash indicator.
Fig.~\ref{fig:compare_model} reports MAE (a), RMSE (b), and $R^2$ (c) for \model and both baselines, across all four topology--configuration combinations; lower MAE/RMSE and higher $R^2$ indicate a more accurate forecast.

\model outperforms both baselines in every combination, on all three metrics. The largest relative gain is in Non-shared/Dynamic, where \model reduces MAE by 66.4\% relative to Mijumbi et al. and 63.6\% relative to Zhang et al., with $R^2$ improving from approximately 0.90 for both baselines to 0.981; under Static, the reduction is smaller but still substantial (45.1\% and 42.7\%). Shared-VNF follows the same pattern, with \model reducing MAE by 48--65\% relative to the two baselines across both configurations.

This gap follows directly from the two architectural differences: neither baseline encodes Intra-VNF metric correlations before graph propagation, and neither conditions message-passing on a crash indicator, so neither can suppress a corrupted signal from a crashing neighbor, the mechanism Section~\ref{sec:crash_impact} shows benefits high-crash-rate VNFs specifically. \model's consistent advantage over both indicates that topology-aware forecasting without Intra-VNF and crash-aware modeling, the design shared by both baselines, leaves a measurable amount of predictable error unaddressed in a realistic, crash-prone SFC deployment.

\vspace{-15pt}

\subsection{Robustness and Deployability Across Operating Conditions}
\label{sec:eval_deployability}

Proactive SFC orchestration requires that forecast accuracy holds up at the longer horizons an orchestrator needs for lead time, and that the model remains computationally viable in production. This subsection evaluates both.
          
\subsubsection{Computational Complexity}
\label{sec:computational_complexity}

\begin{figure}[htbp]
\vspace{-10pt}
    \centering
    \includegraphics[scale=0.49]{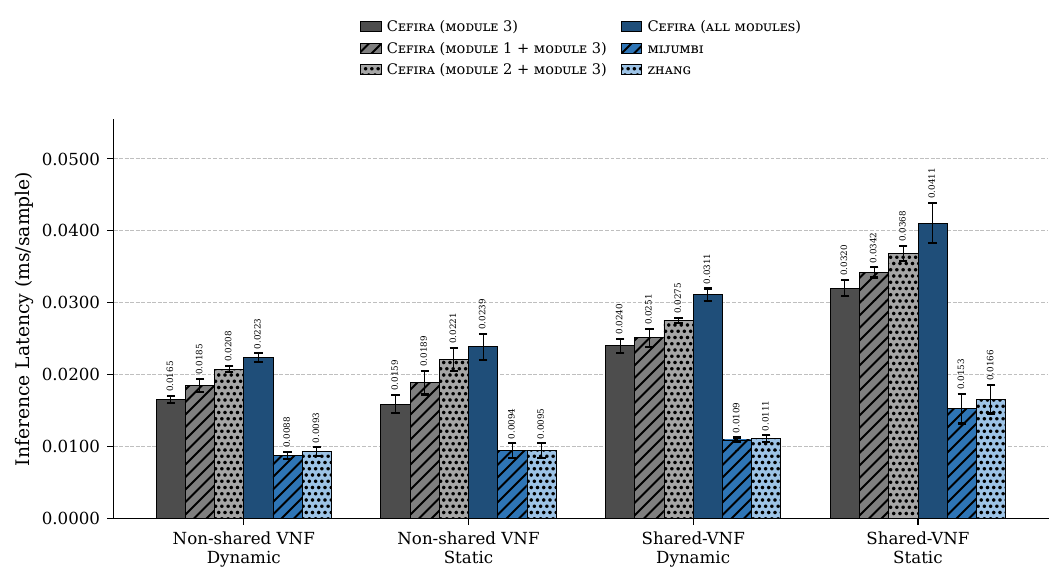}
    \caption{Inference latency per sample of \model and baseline models across topologies and computing resource configurations.}
    \label{fig:inference}
    \vspace{-5pt}
\end{figure}

Fig.~\ref{fig:inference} reports the average inference latency per sample, with standard deviation over 50 independent runs, across all four topology--configuration combinations.

\model achieves 0.022--0.041\,ms per sample, roughly six orders of magnitude below the 30-second monitoring interval (Section~\ref{sec:monitoring_implementation}), imposing no constraint on real-time orchestration. The incremental cost of each module is consistent across topologies: Module~1 adds 5--19\% over Module~3 alone, and Module~2 adds a further 15--39\%, together accounting for the full \model overhead of 28--50\% over Module~3 alone. This overhead is attributable to Module~1's per-VNF intra-metric encoders, Module~2's crash-aware message passing, and Module~3's four temporal backbones. Both baselines achieve lower inference latency (0.009--0.017\,ms per sample), but \model reduces MAE by 45--66\% relative to them (Section~\ref{sec:compare_model}), as neither baseline incorporates intra-VNF metric encoding, temporal sequence modeling, or crash-aware message passing.

\subsubsection{Multi-Horizon Prediction}
\label{sec:multi_horizon}

\begin{figure}[htbp]
\vspace{-10pt}
    \centering
    \includegraphics[scale=0.48]{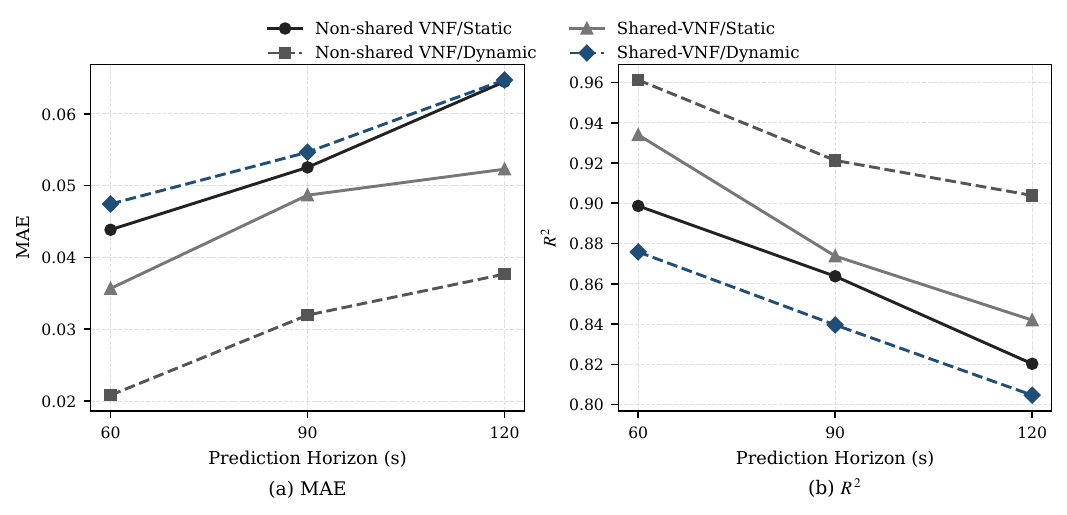}
    \caption{Multi-horizon prediction accuracy across four topology--configuration combinations: (a) MAE and (b) $R^2$.}
    \label{fig:multi-horizon}
    \vspace{-10pt}
\end{figure}

A longer forecast horizon gives an orchestrator more lead time to act, at the cost of a harder forecasting problem; which horizon is preferable depends on how quickly an operator's own scaling or placement pipeline can act once a forecast is available~\cite{lange2019predicting}. Fig.~\ref{fig:multi-horizon} reports MAE (a) and $R^2$ (b) as the horizon extends from 30\,s to 120\,s, for all four combinations.

Accuracy decreases monotonically with horizon: averaged across configurations, $R^2$ falls from approximately 0.96 at 30\,s to 0.84 at 120\,s. The decline is smallest for Non-shared/Dynamic ($R^2\approx0.90$ at 120\,s) and largest for Shared-VNF/Dynamic ($R^2\approx0.80$), consistent with that combination also being the hardest at 30\,s in Sections~\ref{sec:eval_three_modules}--\ref{sec:per_performance_metric}. This follows from \model's delta-mapping (Eq.~\eqref{eq:delta_mapping}): the meta-learner predicts a displacement from the last observed value $\mathbf{x}_i(t)$, and at a longer horizon that observation is a weaker predictor, so the residual grows. \model retains $R^2$ above 0.80 in every combination even at 120\,s, indicating it remains usable well beyond the 30-second horizon used elsewhere in this evaluation; an operator can trade some accuracy for lead time within this range, weighed against their own pipeline's reaction time.

\section{Conclusion}
\label{sec:conclusion}

This paper presented \model, a modular VNF performance forecasting framework that jointly addresses three limitations of existing approaches: intra-VNF metric correlations, inter-VNF topological dependencies, and crash-aware signal propagation across SFCs. Evaluated on a cloud-native testbed spanning two SFC topologies and two resource configurations, \model achieves an average $R^2$ of up to 0.98, outperforms two topology-aware baselines in every setting, and retains $R^2 {>} 0.80$ at a 120-second forecast horizon, well within the operational requirements of proactive NFV orchestration. 
The crash indicator alone reduces MAE by up to 22.4\% at high-crash-rate VNFs, 
confirming that failure-awareness is a necessary component of accurate SFC forecasting. 
Future work will use \model's forecasts as input to a proactive autoscaling system to evaluate end-to-end resource management across larger SFC deployments.

\bibliographystyle{ieeetr}
\bibliography{Refs.bib}

\end{document}